\documentclass[runningheads]{llncs}
\pdfoutput=1
 
\usepackage{eccv}

\usepackage{eccvabbrv}

\usepackage{graphicx}
\usepackage{booktabs}

\usepackage{amssymb, amsmath} 
\usepackage{subfiles}
\usepackage{multirow}
\usepackage{enumitem}
\usepackage{array}
\newcolumntype{I}{!{\vrule width 1pt}}
\usepackage{subcaption}

\setlist[itemize]{label=\textbullet}

\usepackage[accsupp]{axessibility}  

\usepackage{hyperref}

\usepackage{orcidlink}

\begin{document}


\title{MacDiff: Unified Skeleton Modeling with \\Masked Conditional Diffusion} 

\titlerunning{Unified Skeleton Modeling with Masked Conditional Diffusion}

\author{Lehong Wu\inst{1,2}\and
Lilang Lin\inst{1} \and
Jiahang Zhang \inst{1} \and
Yiyang Ma \inst{1} \and
Jiaying Liu\inst{1}\thanks{Corresponding author.}\orcidlink{0000-0002-0468-9576}
}

\authorrunning{L.~Wu et al.}

\institute{
Wangxuan Institute of Computer Technology, Peking University \and
School of Electronics Engineering and Computer Science, Peking University \\
\email{aladonwlh@stu.pku.edu.cn \\ \{linlilang, zjh2020, myy12769, liujiaying\}@pku.edu.cn}
}

\maketitle

\begin{abstract}
Self-supervised learning has proved effective for skeleton-based human action understanding. 
However, previous works either rely on contrastive learning that suffers false negative problems or are based on reconstruction that learns too much unessential low-level clues, leading to limited representations for downstream tasks.
Recently, great advances have been made in generative learning, which is naturally a challenging yet meaningful pretext task to model the general underlying data distributions.
However, the representation learning capacity of generative models is under-explored, especially for the skeletons with spacial sparsity and temporal redundancy.
To this end, we propose Masked Conditional Diffusion (MacDiff) as a unified framework for human skeleton modeling. 
For the first time, we leverage diffusion models as effective skeleton representation learners.
Specifically, we train a diffusion decoder conditioned on the representations extracted by a semantic encoder. Random masking is applied to encoder inputs to introduce a information bottleneck and remove redundancy of skeletons.
Furthermore, we theoretically demonstrate that our generative objective involves the contrastive learning objective which aligns the masked and noisy views. Meanwhile, it also enforces the representation to complement for the noisy view, leading to better generalization performance. 
MacDiff achieves state-of-the-art performance on representation learning benchmarks while maintaining the competence for generative tasks. 
Moreover, we leverage the diffusion model for data augmentation, significantly enhancing the fine-tuning performance in scenarios with scarce labeled data.
Our project is available at \url{https://lehongwu.github.io/ECCV24MacDiff/}.
\end{abstract}

\section{Introduction}
\label{sec:intro}
Human action understanding has been a crucial problem in computer vision. Skeletons use 3D coordinates to represent human joints, providing a lightweight, compact, and privacy-preserving data modality of human representation. Owing to these advantages, skeletons are widely used for human action analysis in real-world applications such as human-robotics interaction \cite{lee20robot}, autonomous driving \cite{camara20pedestrian} and video surveillance \cite{Flaborea23mocodad}. Since well-annotated data are expensive to obtain, self-supervised methods have been proposed to extract meaningful representations from unlabeled skeletons.

Prevalent high-performing self-supervised methods for skeletons mainly contain two paradigms, \ie, contrastive learning-based methods and reconstruction-based methods. Contrastive learning (CL) trains the model to capture shared information between two augmented views of the same sample. Numerous works~\cite{Rao21ascal,Thoker21isc,Chen22skelemixclr,Guo22aimclr,Lin23actclr} focus on designing skeleton-specific augmentations to benefit CL performance. Reconstruction-based methods~\cite{MAE,VideoMAE,Wu22skeletonmae,MAMP} carefully design reconstruction targets that prompt the model to capture the spatial-temporal correlations of skeletons.
However, existing self-supervised methods all focus on specific semantics required by the pre-defined tasks. For example, contrastive learning is proved to only learn the discriminative information for positive/negative pairs~\cite{Wang22Rethinking}, which limits its generalization ability and makes it sensitive to the augmentation design. Reconstruction-based methods overly focus on full signal reconstruction, causing the representations to contain too much low-level information irrelevant for high-level action understanding. 

Generative models, aiming to approximate the real-world data distribution, 
naturally form a more general self-supervised learning target than carefully designed tasks.
Thus, introducing generative models in self-supervised learning would force the model extract better representations with richer semantics.
Autoencoder-based generative models like VAEs~\cite{VAE,VQVAE} rely on the information bottleneck to obtain a compact meaningful latent representation, resulting in a trade-off between discriminability and generation authenticity. More recently, diffusion models~\cite{DDPM,ADM,SD,imagen} have shown remarkable generative capacities. However, diffusion models which directly predict the noise contained in the noisy states do not explicitly learn a meaningful latent representation tailored for discriminative tasks. 
Moreover, due to the spatial sparsity and temporal redundancy of skeletons, diffusion models on skeletons maintain too much irrelevant low-level appearance information, which fails to construct a tight bottleneck and obtain meaningful representation.
Therefore, efforts should be made to mine the potential of such powerful diffusion models to obtain more powerful representations.

To tackle these challenges, we propose Masked Conditional Diffusion (MacDiff) which, for the first time, tames diffusion for both skeleton representation learning and skeleton generation. 
Specifically, we train a semantic encoder to guide a diffusion decoder. The encoder-decoder design serves to disentangle the high-level representation learning with low-level generative training. 
Meanwhile, this architecture mitigates the conflict between discrimination and generation authenticity, allowing for a more flexible design of the encoder.
Considering the spatial-temporal correlation of skeletons, a tighter information bottleneck should be imposed in addition to restricting the dimensions of representations. To this end, we propose to apply random masking on patchified skeletons with a high masking ratio to constrain the representation's dimension.

We conduct theoretical analyses on MacDiff from a mutual information perspective to prove the effectiveness of our framework. We demonstrate that the training of MacDiff is equivalent to a combination of the contrastive learning and reconstruction objective. The contrastive learning objective, conducted on the masked and noisy views of skeletons, enforces the representation to capture shared information between different views, while the very small portion of reconstruction objective enriches the representation by complementing the missing information in the noisy view, including more downstream task-relevant information in the representation.

We provide thorough experiments and results on NTU RGB+D \cite{NTU60,NTU120} and \\PKUMMD \cite{PKUMMD} datasets to demonstrate the effectiveness and versatility of our method. MacDiff achieves remarkable results on self-supervised learning benchmarks. 
In addition, we utilize the pre-trained diffusion for generating label-preserving training data in scenarios with limited training data, which brings significant performance gain in semi-supervised protocols.
Our contributions can be summarized as follows:
\begin{itemize}
\item We propose Masked Conditional Diffusion (MacDiff), a unified framework for human skeleton modeling. A semantic encoder is employed, learning  high-level compact representations, to assist the conditioned generative learning of the diffusion decoder. By virtue of this, our model learns powerful representations for both discriminative and generative downstream tasks. 
\item We theoretically demonstrate that the generative objective of MacDiff involves both diffusion learning and contrastive learning that aligns the representation of the masked view with the noisy view. Moreover, MacDiff is capable of preserving more semantics in the learned representation, leading to better downstream performance than contrastive-only paradigms.
\item MacDiff achieves state-of-the-art performance on three large-scale benchmarks. Remarkably, we leverage diffusion-based data augmentation for encoder fine-tuning, significantly improving the action recognition performance with scarce labeled data.

\end{itemize}


\newcommand{\norm}[1]{\lVert\, #1 \,\rVert}
\newcommand{\bx}{\boldsymbol{x}}
\newcommand{\bz}{\boldsymbol{z}}
\newcommand{\by}{\boldsymbol{y}}
\newcommand{\bX}{\boldsymbol{X}}
\newcommand{\bZ}{\boldsymbol{Z}}
\newcommand{\bY}{\boldsymbol{Y}}
\newcommand{\bI}{\boldsymbol{I}}
\newcommand{\bE}{\boldsymbol{E}}
\newcommand{\normaldistribution}{\mathcal{N}(\boldsymbol{0},\bI)}
\newcommand{\smallheading}[1]{\noindent\textbf{#1.}\hspace{1mm}}
\section{Related Work}

\smallheading{Self-Supervised Learning for Skeletons} Self-supervised learning aims to extract meaningful representations from unlabeled data to facilitate downstream tasks. 
Prevalent methods for skeleton representation learning can be divided into contrastive methods and reconstruction methods. Contrastive learning (CL) extracts meaningful representations by discriminating positive/negative sample pairs from different augmented views \cite{simclr,moco}. To leverage CL for skeletons, numerous works \cite{Rao21ascal,Thoker21isc,Guo22aimclr,Lin23actclr} focus on developing skeleton-specific data augmentations. 
Other works extract shared information between different skeleton modalities \cite{Li21crossclr,Mao23cmd}. Most of these methods use RNNs or GCNs as backbone.
Among reconstruction methods, LongT GAN \cite{Zheng18longtgan} and P\&C \cite{Su19PandC} design reconstruction tasks on autoencoders to learn compressed representations. GL-Transformer \cite{Kin22gltrans} designs prediction tasks with Transformer backbone to capture spatial-temporal dynamics of skeletons. Some works combine reconstruction or prediction tasks with contrastive learning, including \cite{Chen21hitrs,Lin20ms2l,PCM3}. More recently, Masked Autoencoders (MAEs) are introduced to skeletons by SkeletonMAE~\cite{Wu22skeletonmae} and MAMP \cite{MAMP}, which achieve remarkable performance by modeling spatial-temporal correlations of skeletons with Transformers.

\smallheading{Diffusion Models} Diffusion models are a type of generative model that gradually maps a noise prior to a target distribution. They have demonstrated superior capacity in image generation \cite{DDPM,ADM,SD,imagen} and a variety of applications \cite{sdedit,repaint}. Therefore, diffusion models have been widely used for text-guided human motion generation \cite{MDM,Chen23MLD,Gong23diffpose,Zhang24motiondiffuse}, motion prediction \cite{Chen23humanmac} and anomaly detection \cite{Flaborea23mocodad}. These efforts adapt image diffusion models to skeletons with modifications including specialized loss functions for skeletons and leveraging Transformer architecture.

Among a few works that pioneer in leveraging diffusion representations, most works directly use intermediate representations of pre-trained diffusion models or fine-tune the model. While proved effective for dense prediction tasks like segmentation \cite{Namekata24emerdiff} and dense matching \cite{Nam24DenseMatching}, these representations are not compact enough compared with other self-supervised methods. Another paradigm \cite{PDAE} is to distill information from fixed diffusion models. More recently, efforts that jointly train a separate encoder to guide diffusion yield promising results \cite{DiffAE,SODA}. 
In this paper, we propose the first framework that tailors diffusion models for skeleton representation learning, to the best of our knowledge. Besides, we provide theoretical analysis for the effectiveness of the proposed framework.

\smallheading{Unified Skeleton Modeling} Recent advancements in Natural Language Processing have demonstrated the potential for building models that unify multiple tasks. For a general-purpose human representation learning, several works \cite{Lin20ms2l,Chen21hitrs,PCM3} leverage multiple tasks as self-supervised training. MotionBERT \cite{MotionBERT} learns unified 2D skeleton representation by training on 2D-to-3D lifting task. UniHCP \cite{Ci23UniHCP} provides a unified model for several human-centric tasks, \eg, pose estimation, ReID and pedestrian detection. UPS \cite{UPS} forms skeletons and action labels as language tokens. Skeleton-in-Context \cite{Wang23Skeleton-in-Context} leverages in-context learning to unify multiple estimation and prediction tasks. However, most unified methods are restrained to the specific tasks they are trained on. GFPose \cite{Ci23gfpose} utilizes score-based generative model to learn a unified human pose prior but only focus on single-frame skeletons, limiting its applications. In this paper, we explore the capability of a versatile generative model, diffusion model, to unify skeleton representation learning with generation.

\section{Method}

\subsection{Diffusion Models Preliminary}
Diffusion models are a family of generative models that learn the target distribution by performing data denoising.
Denoising Diffusion Probabilistic Model (DDPM) \cite{DDPM} employs a forward (diffusion) process that sequentially corrupts the data distribution $q({x}_{0})$ to the standard Gaussian distribution $\normaldistribution$ with the conditional distribution $q({x}_{t}|{x}_{t-1})$. The noise level of the timesteps $t$ is defined by a fixed increasing variance schedule $\{{\beta}_t\}^T_{t=0}$.
This process can be denoted as:
\begin{equation}
q(\bx_t|\bx_{t-1}) = \mathcal{N}(\bx_t;\sqrt{1-\beta_t}\bx_{t-1},\beta_t\bI).
\end{equation}
With a given $t$, the forward process allows sampling $\bx_t$ directly from $\bx_0$ with $q(\bx_t|\bx_0)=\mathcal{N}(\bx_t;\sqrt{\overline{\alpha}_t}\bx_0,(1-\overline{\alpha}_t)I)$, where $\alpha_t = 1 - \beta_t$ and $\overline{\alpha}_t = \prod_{i=0}^t \alpha_i$. 

The reverse process is defined as another Markov Chain parameterized by $\theta$, which maps the standard Gaussian distribution to the distribution of clean data by gradually denoising. Each step is a Gaussian distribution with a predicted mean $\boldsymbol{\mu}_\theta(\bx_t,t)$ and covariance matrix $\sigma_t^2\bI$:
\begin{equation}
p_\theta(\bx_{t-1}|\bx_t) = \mathcal{N}(\boldsymbol{\mu}_\theta(\bx_t,t), \sigma_t^2\bI).
\end{equation}

The final training objective is derived from optimizing the variational bound on $\mathbb{E}[\,-\text{log}\,p_\theta(\bx_0)\,]$ \cite{DDPM}, where $\gamma_{1:T}$ are positive coefficients depending on $\alpha_{1:T}$:
\begin{equation}
\mathcal{L}(\theta) = \mathbb{E}_{\bx_0,t,\epsilon\sim\normaldistribution}
\left[\,
\gamma_t\norm{
\epsilon - \boldsymbol{\epsilon}_\theta(\sqrt{\overline{\alpha}_t}\bx_0+\sqrt{1-\overline{\alpha}_t}\epsilon, t)
}^2
\,\right].
\label{eq:DDPM loss}
\end{equation}
Other implementations beyond $\boldsymbol{\epsilon}$-prediction include $\bx_{t-1}$- and $\bx_0$-prediction, which are all equivalent to predicting $\boldsymbol{\mu}_\theta(\bx_t,t)$ in mathematical formulation.

 For sampling, we adopt Denoising Diffusion Implicit Model (DDIM) \cite{DDIM}, which shares the same training objective with DDPM but defines a non-Markov diffusion process. DDIM allows for deterministic sampling with better quality. 
\subsection{Masked Conditional Diffusion}
\label{subsec: MacDiff}

\begin{figure}[t]
 \centering
  \includegraphics[width=0.99\textwidth, trim=0cm 0cm 0cm 0cm ]{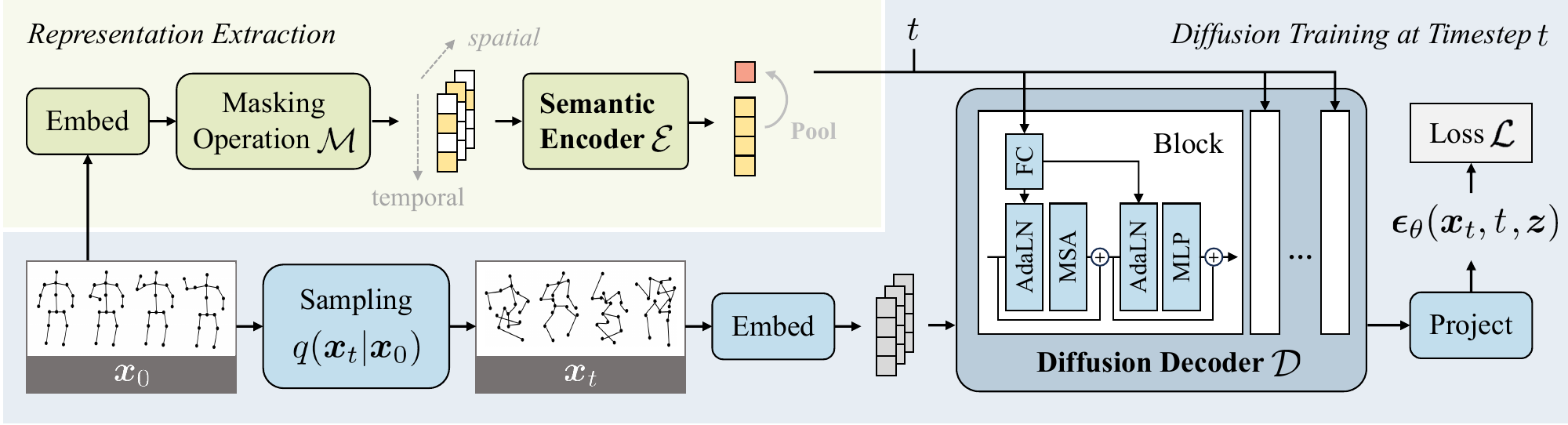}
  \caption{
  The overview of the proposed method. We train a diffusion decoder conditioned on the representations extracted by a semantic encoder.
  In the above stream, we embed the input skeletons into tokens and employ random masking. The global representation is obtained by pooling the local representations extracted by the semantic encoder. In the below stream, we train a conditional diffusion model. We sample the noisy skeleton $\bx_t$ following the diffusion process $q(\bx_t|\bx_0)$. The diffusion decoder predicts the noise $\epsilon$ from $\bx_t$ guided by the learned representation $\bz$. The pre-trained encoder can be utilized independently in downstream discriminative tasks. 
  }
 \label{fig:pipeline}
\end{figure}

In this section, we describe the proposed method, Masked Conditional Diffusion (MacDiff), as a unified framework for human skeleton modeling. \cref{fig:pipeline} illustrates the overall pipeline of our method. 

Concretely, we jointly optimize a semantic encoder $\mathcal{E}$ and a denoising decoder $\mathcal{D}$ (also denoted as $\boldsymbol{\epsilon}_\theta$) by training a conditional diffusion model $\boldsymbol{\epsilon}_\theta(\bx_t, t, \bz)$. 
First, the encoder extracts a latent representation from the masked \textit{view} of skeleton data $\bx$, formulated as $\bz = \mathcal{E}(\mathcal{M}(\bx))$ where $\mathcal{M}(\cdot)$ denotes the masking operation.
Then, the decoder predicts the noise contained in $\bx_t$ conditioned on $\bz$.
With the aforementioned pre-training, the encoder learns to extract compact representation applicable for discriminative downstream tasks.

\smallheading{Patchify and Embedding} 
Given the input skeleton $\bx\in\mathbb{R}^{T_0\times V\times 3}$, we first divide it into non-overlapping patches with equal length along the temporal dimension $\bx'\in\mathbb{R}^{T_p\times V\times (l\times 3)}$. $T_0$ is the number of frames, $V$ is the number of joints, $l$ is the patch length, and $T_p = T_0 / l$. 
The patches are then flattened and embedded to $C$ dimensions with a trainable linear projection:
$\bE = \text{Embedding}(\bx')\in\mathbb{R}^{T_p\times V \times C}$.
Now we have $T\cdot V$ tokens of dimension $C$ as the encoder's input.

\smallheading{Random Masking} Considering the redundancy in the skeleton, especially in the temporal dimension, naive reconstruction training can result in the undesirable shortcuts and degrade the model performance. 
Meanwhile, based on the information bottleneck principle \cite{Goldfeld20IB,Liu22MRCL}, the model learns a compact representation if information compression is introduced to remove redundancy.
Therefore, we employ random masking for the encoder input with a masking ratio $r$, retaining a total of $K =  \left\lceil (1-r) \cdot T_p \cdot V \right\rceil$ tokens.
In practice, we adopt an extremely high masking ratio $r = 90\%$, enforcing a tight bottleneck on the representation.
In addition, the masking operation significantly speeds up training by reducing the encoder's computational graph.

\smallheading{Model Architecture} Our encoder and decoder both follow a vanilla Transformer~\cite{Attention} architecture.
First, trainable spatial and temporal positional embeddings $\bE^{s}_{pos} \in \mathbb{R}^{1 \times V \times C}, \bE^{t}_{pos} \in \mathbb{R}^{T_p \times 1 \times C}$ are added to the tokens $\bE$ with broadcasting.
Note that this is implemented before the masking operation for the encoder.
The Transformer network consists of alternating layers of multi-head self-attention (MSA) and multi-layer perceptron (MLP) with residual connection. Layer Norm (LN) is applied before each layer and after the last layer.  

The encoder output $\bz_{local} \in \mathbb{R}^{K\times C}$ is the local representations corresponding to unmasked patches. The global representation $\bz_{global}$ is obtained by pooling all tokens.
 For the decoder output, we linearly project it to the final prediction of the same shape as $\bx_0$.

\smallheading{Conditioning} To incorporate the condition $\bz$ into the denoising decoder, we replace Layer Norm with Adaptive Layer Norm (AdaLN), following~\cite{DiT}:
\begin{equation}
    \mathrm{AdaLN}(\boldsymbol{h},\bz,t) = \bz_s \cdot (t_s \cdot \mathrm{LN}(\boldsymbol{h}) + t_b) + \bz_b,
\end{equation}
where $\boldsymbol{h}$ is the hidden representation, $(t_s,t_b)$ and $(\bz_s,\bz_b)$ are obtained from linear projection of the timestep embedding $t$ and condition $\bz$, respectively. Through AdaLN layers, the condition $\bz$ guides the denoising process by scaling and shifting normalized hidden representation.
To further disentangle the contributions of the encoder and decoder to the final prediction, we dropout $\bz$ with a probability of 0.1, meanwhile enabling unconditional generation.

In practice, we observe an over-smoothing problem of the encoder if we simply utilize $\bz_{global}$ (with broadcast) as $\bz$. Over-smoothing is a common problem for GCNs and Transformers \cite{Huang20GCNOversmooth, Dong21transformerOversmooth} that degrades performance and can be reflected by high similarity between tokens. We aim to increase token uniformity by preserving more local information in tokens in addition to global information. To this end, we unshuffle the unmasked tokens $\bz_{local}$ and fill the masked positions with $\bz_{global}$ to form $\bz$. Since a high masking ratio is implemented, the model prioritizes optimizing the global representation while also attempts to benefit from local information by directly optimizing local representations. 

\smallheading{Diffusion Training} Considering the discrepancy between the statistics of original skeleton data and the standard Gaussian distribution, we normalize the data using the mean $\mu$ and standard variation $\sigma$ calculated from the training set: 
\begin{equation}
    \bx_0 = \frac{\bx_{orig} - \mu}{\sigma}, \qquad
    \mu,\sigma \in \mathbb{R}^{1 \times 1 \times 3}.
\end{equation}

For detailed settings of diffusion, we adopt some common practices including $\boldsymbol{\epsilon}$-prediction, total timesteps $T = 1000$ and loss weights $\gamma_t = 1$ \cite{DDPM}. Our training objective is a simplified and conditional version of \cref{eq:DDPM loss}:
\begin{equation}
\mathcal{L} = 
\mathbb{E}_{\bx_0,t,\epsilon}
\left[\,
\norm{
\epsilon - \mathcal{D}(\sqrt{\overline{\alpha}_t}\bx_0+\sqrt{1-\overline{\alpha}_t}\epsilon \,,\, t
 \,,\, \mathcal{E}(\mathcal{M}(\bx_0))
)}^2
\,\right].
\end{equation}

A minority choice in our work regarding diffusion is the inverse-cosine schedule \cite{SODA}. The inverse-cosine schedule pulls the noise level of all timesteps towards a medium level compared with commonly-used cosine \cite{Nichol21improvediffusion} or linear \cite{DDPM} schedules. We verify this choice with experiments (see \cref{ablation}).

\subsection{Information Analysis on MacDiff}
\label{subsec: Theory}
In this section, we conduct information-theoretic analyses on our proposed framework MacDiff. We formulate the generative objective of MacDiff as an improvement of contrastive learning (CL) objectives, leading to a better guarantee of downstream performance.
For mathematical formulation, we use random variables $X, X_t, X_m$ and $Z$ to denote the original view, noisy view, masked view, and latent representation of the skeleton data. 

\smallheading{The Training Objective of MacDiff} A generative model (\eg, VAEs and MAEs) that predicts some target $X$ from latent code $V$ maximizes their mutual information (MI), \ie $I(X;V)$. In the case of MacDiff, $V$ takes the form of $(Z, X_t)$. MacDiff can thereby be described as: 
\begin{equation}
\label{eq: MI1}
    \underset{\mathcal{E},\mathcal{D}}{\text{max}} \;\; 
    I(X;(Z, X_t)), \quad Z = \mathcal{E}(X_m).
\end{equation}
This training objective is further formulated as follows:
\begin{equation}
\label{eq: MI2}
 I(X;(Z, X_t)) = I(X;Z) + I(X;X_t|Z).
\end{equation}
The first term $I(X;Z)$ trains the encoder to contain more information about $X$ in representation $Z$, while the second term trains the decoder to predict $X$ from $X_t$ conditioned on $Z$. By employing the encoder-decoder design and dropouting on the representation, we further disentangle the encoder's high-level representation learning from the decoder's low-level prediction objective.

\smallheading{MacDiff as an Improvement of Contrastive Learning}
We can further decompose $I(X;Z)$ into two terms with the intermediate variable $X_t$:
\begin{equation}
\label{eq: MI3}
    I(X;Z) = I(X_t;Z) + I(X;Z|X_t).
\end{equation}
 We point out that the first term is consistent with the contrastive learning objective. 
CL assumes that the information needed for downstream task $Y$ is shared between two views $X_1,X_2$\cite{Wang20InfoMin,Ziv23SSLMI,Wang22Rethinking}. Therefore, CL aims to optimize the MI between their representations $Z_1, Z_2$. 
\begin{equation}
    I(Z_1;Z_2) \leq I(Z_1;X_2) \leq I(X_1;X_2) = I(X;Y).
\end{equation}
In our case, $X_1,X_2$ correspond to the masked and noisy views $X_m, X_t$. MacDiff does not directly extract representation from $X_t$. It instead optimizes $I(Z_1, X_2) = I(Z, X_t)$ as a tighter lower bound of $I(X_1;X_2)$, which avoids feature collapse.

Moreover, the second term aims to contain more information about $X$ in $Z$ that is complementary to $X_t$. CL is proved to suffer from the \textit{discriminative information overfitting} problem \cite{Liu22MRCL} that the model is biased to extract only the discriminative information of two views.
However, optimizing $I(X;Z|X_t)$ requires the representation to contain more task-relevant information that is not shared between views. Therefore, the generative objective of MacDiff can be viewed as an improvement of CL methods and provides a better theoretical guarantee of downstream performance as discussed next.

\smallheading{Relation to Downstream Performance} The Bayes error rate $P_e$ is the lowest error that can be achieved by any classifier trained on the given data representations, defined as $P_e = 1 - \mathbb{E}_{z\sim p(Z)}
[\; \text{max}_{y\in Y} \; p(y|z) \;]$, where $Z$ denotes the representation and $Y$ denotes the labels. Then, we can prove that (refer to the supplementary material):
\begin{theorem}
(Bayes Error Rate of Representations) For arbitrary data representation distribution $Z$, and $V$ denotes a certain view of the data, its Bayes error rate can be estimated as:
\begin{align}
    P_e &\leq 1 - e^{-(H(Y) - I(Z;Y))} \\
        &\leq 1 - e^{-(H(Y) - I(Z;Y;V) - I(Z;Y|V))}.
\end{align}
\end{theorem}
We set $V = X_t$ in this theorem. Thus, our goal is to increase the terms $I(Z;Y;X_t)$ and $I(Z; Y | X_t)$. 
These two terms are bounded by $I(Z; X_t)$ and $I(Z; X | X_t)$ respectively. As shown in \cref{eq: MI3}, the MacDiff objective directly increases both terms. Note that the contrastive learning objective merely optimizes the first term $I(Z; X_t)$, which explains the improved downstream performance of our method compared to contrastive-only methods.

\subsection{Diffusion-Based Data Augmentation}
For generative self-supervised methods (\eg, MAEs), only the encoders are utilized for downstream tasks, while the rest of the models are completely discarded. In MacDiff, we propose that the denoising decoder can be used for data augmentation when fine-tuning the encoder. We focus on scenarios where labeled data is scarce, which is quite common in reality, especially for skeletons.

To synthesize training data that follows the real-data distribution $q(\bx,y)$, generative methods have to be label-preserving. Existing methods achieve this by leveraging pre-trained text-guided diffusion \cite{Zhang23expand,Yuan24realfake}. However, our semi-supervised setting does not provide text guidance, which forms a more general challenge of synthesizing label-preserving samples given labeled ones. 

To this end, we propose diffusion-based data augmentation based on the assumption that samples generated with the same representation guidance are label-consistent. Specifically, we take two steps: (1) pre-calculate the representations of labeled samples, and (2) generate new samples with the decoder conditioned on these representations. Nevertheless, sampling from Gaussian noise is time-consuming and thus can only be performed before training. We find that one-step denoising from some medium timestep $t_s$ yields similar effects with smaller computational cost, which allows for generating diverse augmented data at different epochs. 

\section{Experiments}

\subsection{Experimental Setup}
\smallheading{Datasets} For evaluation, our experiments are conducted on the following three datasets: NTU RGB+D 60 dataset (NTU 60)~\cite{NTU60}, NTU RGB+D 120 dataset (NTU 120)~\cite{NTU120} and PKU Multi-Modality Dataset (PKUMMD)~\cite{PKUMMD}. All three datasets use 25 joints to represent the human body.

NTU 60 is a large-scale dataset for human action recognition with 60 categories and 56,578 videos. We follow the widely-used evaluation protocols, cross-subject (xsub) and cross-view (xview). The former uses action sequences from half of the 40 subjects for training, and the rest for testing. The latter uses sequences from camera 2,3 for training and sequences from camera 1 for testing. 

NTU 120 is an extension of NTU 60, with 120 categories and 114,480 videos from 106 subjects. Evaluation protocols on NTU 120 are cross-subject (xsub) and cross-setup (xset). Specifically, xset divides sequences into 32 setups based on the camera distance and background, half of which are used for training and the rest for testing. 

PKUMMD covers a multi-modality 3D understanding of human actions, with 52 categories and almost 20,000 instances. PKUMMD is divided into part I and II, and part II is more challenging due to the noise caused by view variation. We split training and testing sets according to the cross-subject protocol. 

\smallheading{Implementation Details}  The input sequence of 300 frames is cropped and interpolated to 120 frames, and the patch length $l=4$. Apart from random crop, we use random rotation and small Gaussian noise ($\sigma=0.005$) as data augmentation. Note that the small noise is only added to encoder inputs. 
For Transformer architecture, the embedding dimension is 256, the MLP hidden dimension is 1024, and the number of heads in MSA is 8. By default, the encoder and decoder have 8 layers and 5 layers, respectively.
We train our model on four NVIDIA TITAN Xp GPUs with a total batch size of 128 for 500 epochs. The AdamW optimizer is adopted with the learning rate decreasing from 1e-3 to 1e-5.

\subsection{Self-Supervised Learning Evaluation}

\begin{table}[t]
\scriptsize
  \centering
  \caption{Comparison of linear evaluation results on NTU 60, NTU 120, and PKUMMD datasets. 3s- represents the ensemble results of joint(J), bone(B) and motion(M) streams. \textbf{Bold} and \underline{underlined} indicate the best and second best results, respectively. The same notation applies throughout.}
  \setlength{\tabcolsep}{2.0mm}{
  \begin{tabular}{l|c|c|c|c|c|c}
    \toprule
    \multirow{2}{*}{Method} & \multirow{2}{*}{Stream} & \multicolumn{2}{c|}{NTU 60} & \multicolumn{2}{c|}{NTU 120} & \multirow{2}{*}{PKU I}\\
    & & xsub & xview & xsub & xset &   \\
    \midrule
    3s-CrosSCLR~\cite{Li21crossclr}& J+M+B & 77.8 & 83.4 & 67.9 & 66.7 & 84.9 \\
    3s-AimCLR~\cite{Guo22aimclr}& J+M+B & 78.9 & 83.8 & 68.2 & 68.8 & 87.4 \\
    3s-SkeleMixCLR~\cite{Guo22aimclr}& J+M+B & 82.7 & 87.1 & 70.5 & 70.7 & 91.1 \\
    3s-ActCLR~\cite{Lin23actclr}& J+M+B & 84.3 & 88.8 & 74.3 & 75.7 & - \\
    3s-CPM~\cite{Zhang22cpm}& J+M+B & 83.2 & 87.0 & 73.0 & 74.0 & 90.7 \\
    \midrule
    LongT GAN~\cite{Zheng18longtgan}& J & 39.1 & 48.1 & - & - & 67.7 \\
    MS$^{\rm 2}$L~\cite{Lin20ms2l} & J & 52.6 & - & - & - & 64.9 \\
    AS-CAL~\cite{Rao21ascal}& J & 58.5 & 64.8 & 48.6 & 49.2 & - \\
    ISC~\cite{Thoker21isc}& J & 76.3 & 85.2 & 67.1 & 67.9 & 80.9 \\
    GL-Transformer~\cite{Kin22gltrans}& J & 76.3& 83.8 & 66.0 & 68.7 & - \\
    CMD~\cite{Mao23cmd}& J & 79.4 & 86.9 & 70.3 & 71.5 & - \\
    PCM$^{\rm 3}$~\cite{PCM3}& J & 83.9 & \underline{90.4} & 76.5 & 77.5 & - \\
    SkeletonMAE~\cite{Wu22skeletonmae} & J & 74.8 & 77.7 & 72.5 & 73.5 & 82.8 \\
    MAMP~\cite{MAMP} & J & \underline{84.9} & 89.1 & \underline{78.6} & \underline{79.1} & \underline{92.2}  \\
    \textbf{MacDiff (Ours)} & J & \textbf{86.4} & \textbf{91.0} & \textbf{79.4} & \textbf{80.2} & \textbf{92.8} \\
    \bottomrule
  \end{tabular}
   \label{table: linear}
  }
\end{table}
\vspace{-7pt}

\begin{table}[t]

\scriptsize
\noindent
\begin{minipage}[t]{0.48\textwidth}

    \centering
    \captionof{table}{Comparison of supervised fine-tuning evaluation results on NTU 60 xsub and xview datasets. The bottom four rows are unified models, among which UPS is a supervised method.}
    
  \setlength{\tabcolsep}{0.9mm}{
  \begin{tabular}{l|c|c|c}
    \toprule
    \multirow{2}{*}{Method} & \multirow{2}{*}{Backbone} & \multicolumn{2}{c}{NTU 60} \\
    & & xsub & xview  \\
    \midrule
    CrosSCLR~\cite{Li21crossclr} & 3s-ST-GCN & 86.2 & 92.5 \\
    AimCLR~\cite{Guo22aimclr} & 3s-ST-GCN & 86.9 & 92.8 \\
    ActCLR~\cite{Lin23actclr} & 3s-ST-GCN & 88.2 & 93.9 \\
    MCC~\cite{Su21mcc} & 2s-AGCN & 89.7 & 96.3 \\
    SkeletonMAE~\cite{Wu22skeletonmae} & Transformer & 88.5 & 94.7 \\
    MAMP~\cite{Wu22skeletonmae} & Transformer & \textbf{93.1} & \textbf{97.5} \\
    \midrule
    Hi-TRS~\cite{Chen21hitrs} & 3s-Transformer & 90.0 & 95.7 \\
    MotionBERT~\cite{MotionBERT} & DSTformer & \underline{93.0} & 97.2 \\
    UPS~\cite{UPS} & Transformer & 92.6 & 97.0 \\
    \textbf{MacDiff (Ours)} & Transformer & 92.7 & \underline{97.3} \\
    \bottomrule
  \end{tabular}
  \label{table: finetune}
  }
\end{minipage}%
\hfill
\begin{minipage}[t]{0.48\textwidth}
    \centering
    \captionof{table}{Comparison of transfer learning results on PKUMMD II dataset. The source datasets are NTU 60, NTU 120 and PKUMMD I. All datasets use the xsub split.}
    
  \setlength{\tabcolsep}{0.9mm}{
  \begin{tabular}{l|c|c|c}
    \toprule
    \multirow{2}{*}{Method} & \multicolumn{3}{c}{To PKU II} \\
    & NTU 60 & NTU 120 & PKU I\\
    \midrule
    LongT GAN~\cite{Zheng18longtgan} & 44.8 & - & 43.6 \\
    MS$^{\rm 2}$L~\cite{Lin20ms2l} & 45.8 & - & 44.1 \\
    ISC~\cite{Thoker21isc} & 51.1 & 52.3 & 45.1 \\
    CMD~\cite{Mao23cmd} & 56.0 & 57.0 & - \\
    SkeletonMAE~\cite{Wu22skeletonmae} & 58.4 & 61.0 & 62.5 \\
    MAMP~\cite{MAMP} & \underline{70.6} & \underline{73.2} & \underline{70.1} \\
    \textbf{MacDiff (Ours)} & \textbf{72.2} & \textbf{73.4} & \textbf{71.4} \\
    \bottomrule
  \end{tabular}
  \label{table: transfer}
  }
\end{minipage}

\end{table}

\smallheading{Linear Evaluation} In the linear evaluation protocol, a linear classifier is post-attached to the encoder to classify the learned representations. We fix the encoder and train the classifier for 100 epochs with the SGD optimizer and a learning rate of 0.1.  
We compare MacDiff with latest methods, with action recognition accuracy reported as a measurement. 
 
As shown in \cref{table: linear}, our method surpasses high-performing reconstruction-based methods, \eg, SkeletonMAE~\cite{Wu22skeletonmae} and MAMP~\cite{Wu22skeletonmae}. With only the joint stream, our method also outperforms multi-stream contrastive learning methods, \eg 3s-AimCLR~\cite{Guo22aimclr}, 3s-CMD~\cite{Mao23cmd} and 3s-ActCLR~\cite{Lin23actclr}. The result demonstrates that MacDiff captures the spatial-temporal correlation of skeletons better than existing methods, and also confirms our theoretical analysis that MacDiff provides a better framework than contrastive-only paradigms.

\smallheading{Supervised Fine-tuning Evaluation} In the fine-tuning evaluation protocol, we attach an MLP head to the pre-trained encoder and train the whole model for another 100 epochs with the AdamW optimizer and the learning rate decreasing from 3e-4 to 1e-5. 

As shown in \cref{table: finetune}, our method yields comparable results to MAMP and outperforms other existing methods. We point out that our performance gap in the fine-tuning protocol with MAMP is trivial since we share the same encoder architecture and the learned representation is disrupted during fine-tuning. Meanwhile, MacDiff outperforms other unified models such as Hi-TRS~\cite{Chen21hitrs}, MotionBERT~\cite{MotionBERT} and UPS~\cite{UPS}.

\smallheading{Transfer Learning Evaluation} In the transfer learning evaluation protocol, the backbones are pre-trained on a source dataset and evaluated on a target dataset following linear evaluation protocol to keep the backbone intact. 

As shown in \cref{table: finetune}, our method achieves significant transfer learning performance on the challenging PKUMMD II, demonstrating the generalization ability and robustness of our method.

\subsection{Semi-Supervised Fine-tuning with Diffusion-based Data Augmentation} 
\label{subsec: augmentation}

\setlength{\textfloatsep}{10pt plus 2pt minus 4pt}
\begin{table}[t]
  \scriptsize
  \centering
  \caption{Comparison of semi-supervised fine-tuning results on NTU 60 datasets. By default we set the starting timestep $t_s=500$ and augment-to-real ratio $\lambda=2.0, 0.5, 0.25$ for $p=1\%, 2\%, 10\%$. The results are averaged over 5 runs. For MacDiff, results with and without diffusion-based augmentation are both reported.}
  \setlength{\tabcolsep}{3.0mm}{
  \begin{tabular}{l|c|c|c|c|c|c}
    \toprule
    \multirow{2}{*}{Method} & \multicolumn{3}{c|}{NTU 60 xsub} & \multicolumn{3}{c}{NTU 60 xview} \\
    & 1\% & 2\% & 10\% & 1\% & 2\% & 10\% \\
    \midrule
    ISC~\cite{Thoker21isc} & 35.7 & - & 65.1 & 38.1 & - & 72.5 \\
    3s-AimCLR~\cite{Guo22aimclr} & 54.8 & - & 78.2 & 54.3 & - & 81.6 \\
    3s-CMD~\cite{Mao23cmd} & 55.6 & - & 79.0 & 55.5 & - & 82.4 \\
    CPM~\cite{Zhang22cpm} & 56.7 & - & 73.0 & 57.5 & - & 77.1 \\
    PCM$^{\rm 3}$~\cite{PCM3} & 53.8 & - & 77.1 & 53.1 & - & 82.8 \\
    SkeletonMAE~\cite{Wu22skeletonmae} & 54.4 & - & 80.6 & 54.6 & - & 83.5 \\
    MAMP~\cite{MAMP} & \underline{66.0} & 80.3 & 88.0 & 68.7 & 83.5 & 91.5 \\
    \midrule
    \textbf{MacDiff w/o aug} & 65.6 & \underline{80.7} & \underline{88.2} & \underline{77.3} & \underline{84.1} & \underline{92.5} \\
    \textbf{MacDiff} & \textbf{72.0} & \textbf{82.1} & \textbf{89.2} & \textbf{79.2} & \textbf{85.6} & \textbf{93.1}\\
    \bottomrule
  \end{tabular}
  \label{table: semi}
  }
  
\end{table}

We next evaluate the effectiveness of the diffusion-based data augmentation in scenarios with limited labeled data. We report results following the semi-supervised protocol, which is consistent with the fine-tuning protocol except that only a proportion $p$ of the training set is used. We evaluate our methods when $p = 1\%, 2\%, 10\%$. 

We explore different augment-to-real ratios $\lambda$, defined as the ratio of augmented data to real data, for different proportions of training data. 
We empirically find that the optimal augment-to-real ratio $\lambda$ declines as $p$ increases. Intuitively, when labeled data is severely scarce, augmented data that falls in the neighborhood of labeled data representations helps the classifier to learn more robust boundaries. However, as the cardinality of the training set increases, since the dataset inherently contains ambiguous class boundaries (\eg, "phone call" and "play with phone/tablet" in NTU 60), generating new data around them further confuse the classifier. See more experiments in \cref{ablation}.

As shown in \cref{table: semi}, with the aid of augmented training samples, MacDiff outperforms state-of-the-art MAMP by 6.0\%, 1.8\%, 1.2\% in NTU 60 xsub, and 10.5\%, 2.1\%, 1.6\% in NTU 60 xview within the three settings, respectively. Compared with encoder-only MacDiff, the augmentation brings significant performance gain of 6.4\%, 1.4\%, 1.0\% in NTU 60 xsub, and 1.9\%, 1.5\% , 0.6\% in NTU 60 xview. 

\subsection{Generative Evaluation}
In this section, we implement MacDiff for motion reconstruction and motion generation tasks. We compare our method with reconstruction-based method SkeletonMAE~\cite{Wu22skeletonmae} and diffusion-based methods DDIM\cite{DDIM} and MDM\cite{MDM}. Note that MAMP\cite{MAMP} cannot be applied for either task because it predicts the normalized motion.
For fair comparison, all methods are implemented with our Transformer decoder architecture. In addition, we also implement the original MDM (denoted as MDM-orig) with temporal-only attention, 8 layers, and 512 hidden dimensions. Please find implementation details in the supplementary material.  
All experiments are conducted on the testing set of NTU 60 xsub.

\smallheading{Motion Reconstruction} Real-world skeleton data suffer from occlusions, resulting in incomplete sequences. 
We evaluate motion reconstruction in two types of occlusions: (1) random consecutive frames, and (2) a random body part from \{trunk, left arm, right arm, left leg, right leg\}, following the division of previous works. 
We follow a diffusion-based inpainting paradigm \cite{repaint} for DDIM and MacDiff. The MacDiff decoder is fine-tuned for another 100 epochs with the encoder fixed and only global representations.

We report Mean Per Joint Position Error (MPJPE) as our metric. As shown in \cref{table: reconstruction}, MacDiff is capable of recovering incomplete skeletons as a unified framework and surpasses reconstruction-based SkeletonMAE and DDIM.

\begin{table}[t]

  \scriptsize
  \centering
  \caption{Comparison of motion reconstruction results on NTU 60 xsub. MPJPE is reported on four occlusion settings: 10 frames, 20 frames, 40 frames and 1 body part.}
  \setlength{\tabcolsep}{3.0mm}{
  \begin{tabular}{l|c|c|c|c}
    \toprule
    & 10 frames & 20 frames & 40 frames & 1 body part \\
    \midrule
    SkeletonMAE~\cite{Wu22skeletonmae} & 0.191 & 0.221 & 0.255 & 0.319\\
    DDIM~\cite{DDIM} & \underline{0.041} & \textbf{0.087} & \underline{0.205} & \underline{0.251} \\
    \textbf{MacDiff (Ours)} & \textbf{0.033} & \underline{0.089} & \textbf{0.147} & \textbf{0.241}\\
    \bottomrule
  \end{tabular}
  \label{table: reconstruction}
  }
  
\end{table}

\begin{table}[t]
  \scriptsize
  \centering
  \caption{Comparison of motion generation results on NTU 60 xsub. MacDiff is evaluated both with and without fine-tuning.}
  \setlength{\tabcolsep}{3.0mm}{
  \begin{tabular}{l|c|c|c|c|c}
    \toprule
    \multirow{2}{*}{Method} & \multirow{2}{*}{Prediction} & \multirow{2}{*}{FID$\downarrow$} & \multirow{2}{*}{KID$\downarrow$} & \multirow{2}{*}{Diversity$\uparrow$} & Precision$\uparrow$ \\
    & & & & & Recall$\uparrow$\\
    \midrule
    Real data & - & 0.05 & 0.0001 & 2.09 & 0.886, 0.908 \\
    \midrule
    MDM-orig~\cite{MDM} & $x_0$ & 3.49 & 0.0328 & 1.51	& 0.047, 0.030 \\
    MDM~\cite{MDM} & $x_0$ & \textbf{1.06} & \textbf{0.0059} & 1.43 & \textbf{0.828}, 0.070\\
    DDIM~\cite{DDIM} & $\epsilon$ & 1.32 & 0.0082 &	1.89 & \underline{0.576}, \textbf{0.537} \\
    \textbf{MacDiff (Ours)} & $\epsilon$ & 1.52 & 0.0085 & \underline{1.96} & 0.414, 0.289 \\
    \textbf{\;\;\; $+$ fine-tune} & $\epsilon$ & \underline{1.30} & \underline{0.0070} & \textbf{1.97} & 0.460, \underline{0.505} \\
    \bottomrule
  \end{tabular}
  \label{table: generation}
  }
  
\end{table}
\vspace{-2pt}

\smallheading{Motion Generation} We utilize the MacDiff decoder for unconditional motion generation. We report four metrics FID, KID, diversity, and precision/recall. Please refer to the supplementary material for detailed implementation of these metrics. Note that reconstruction-based methods are not capable of unconditional generation. As shown in \cref{table: generation}, MacDiff achieves comparable results with DDIM and MDM.

\smallheading{Qualitative Results} We provide visualization of motion reconstruction, unconditional motion generation and one-step denoising results (for data augmentation) in the supplementary material.

\subsection{Ablation Study}
\label{ablation}
\smallheading{Masking Strategy and Ratio} In \cref{table: masking}, we compare the results of different masking strategies, including random masking, temporal-only masking, tube masking \cite{VideoMAE}, spatial-temporal masking and motion-aware random masking \cite{MAMP}. For tube masking, the tube length is set to 5. For spatial-temporal masking, we keep 8 out of 25 joints and 10 out of 30 temporal patches. For motion-aware masking, we follow the implementation of MAMP. The results show that the simple random masking works best as a spacetime-agnostic masking. We also compare different masking ratios and find a high masking ratio of 90\% works best, which coincides with the findings in the video field.

\begin{table}[t]
\scriptsize
\noindent
\begin{minipage}[t]{0.48\textwidth}
    \centering
    \captionof{table}{Ablation study on the masking strategy and the masking ratio. We report results on NTU 60 xsub under the linear evaluation protocol.}
  \setlength{\tabcolsep}{1.mm}{
  \begin{subtable}[b]{.42\linewidth}
  \begin{tabular}{l|c}
    \toprule
    \multirow{2}{*}{Strategy} & NTU 60 \\
     & xsub \\
    \midrule
    Temporal & 84.1\\
    Tube & 83.1 \\
    Spatial-temporal & 85.3\\
    Random & \textbf{86.4}\\
    Motion-aware & 85.5 \\
    \bottomrule
  \end{tabular}
  \end{subtable}
  \hfill
  \begin{subtable}[b]{.32\linewidth}
  \begin{tabular}{l|c}
    \toprule
      \multirow{2}{*}{Ratio} & NTU 60\\
      & xsub \\
    \midrule
     0 & 79.3\\
     50\% & 82.7 \\
     80\% & 83.8 \\
     90\% & \textbf{86.4}\\
     95\% & 83.6\\
    \bottomrule
  \end{tabular}
  \end{subtable}
\label{table: masking}
  }
\end{minipage}%
\hfill
\begin{minipage}[t]{0.48\textwidth}
    \centering
    \captionof{table}{Ablation study on the noise schedule. We report results on NTU 60 xsub under the linear evaluation protocol.}
  \setlength{\tabcolsep}{2.mm}{
  \begin{tabular}{l|c}
    \toprule
     \multirow{2}{*}{Noise Schedule} & NTU 60\\
     & xsub \\
    \midrule
    $\tau=1.5$ & 85.7 \\
    $\tau=1.0$ (Inverse)& \textbf{86.4} \\
    $\tau=0.5$ & 85.8 \\
    $\tau=-0.5$ & 85.2 \\
    $\tau=-1.0$ (Cosine)& 83.8 \\
    Linear & 83.4 \\
    \bottomrule
  \end{tabular}
  \label{table: noise schedule}
  }
\end{minipage}

\end{table}

\smallheading{Noise Schedule} We construct a series of noise schedules as linear combinations of the inverse-cosine~\cite{SODA} and cosine schedule controlled by $\tau$ (see definition in the supplementary material). $\tau=1$ and $\tau=-1$ represents the inverse-cosine and cosine schedule, respectively. We compare these schedules with the widely used linear schedule. As shown in \cref{table: noise schedule}, cosine-based schedule performs better than the linear schedule, and the performance peaks at $\tau=1$, indicating that medium noise levels are preferred for representation learning. 

\smallheading{Diffusion-based Data Augmentation} For the ablation study of our diffusion-based data augmentation, We compare the effects of different starting timestep $t_s$ and augment-to-real ratio $\lambda$. As shown in \cref{table: augmentation}, the performance gain is highest when $t_s = 500$. Intuitively, an overly large $t_s$ may introduce too much noise since we implement one-step denoising, while an overly small $t_s$ fails to provide sufficient strength of augmentation. For larger proportion of training data, the optimal $\lambda$ is smaller, which is consistent with our analysis in \cref{subsec: augmentation}.

\begin{table}[t]
\scriptsize
\noindent
\begin{minipage}[t]{0.6\textwidth}
    \centering
    \captionof{table}{Ablation study on the starting timestep $t_s$ and augment-to-real ratio $\lambda$ of diffusion-based augmentation. We report results on NTU 60 xsub under the semi-supervised 1\% and 10\% protocol. $t_s=0$ or $\lambda=0$ means results without augmentation. }
  \setlength{\tabcolsep}{3.mm}{
  
  \begin{subtable}[b]{.3\linewidth}
  \begin{tabular}{l|c}
    \toprule
    $t_s$ & semi 1\% \\ 
    \midrule
    0 & 65.6 \\
    100 &  66.4 \\
    300  & 71.4 \\
    500  & \textbf{72.0} \\
    900 &  68.7 \\
    \bottomrule
  \end{tabular}
  \end{subtable}
  \hfill
    \begin{subtable}[b]{.6\linewidth}
      \begin{tabular}{l|c|c}
    \toprule
    $\lambda$ &  semi 1\% & semi 10\% \\ 
    \midrule
    0 & 65.6 & 88.2 \\
    0.25 & 67.8 & 89.1 \\
    0.5  & 68.2 & \textbf{89.2} \\
     1.0  & 70.0 & 88.5 \\
    2.0  & \textbf{72.0} & - \\
    \bottomrule
  \end{tabular}

  \end{subtable}
  
  \label{table: augmentation}
  }
\end{minipage}%
\hfill
\begin{minipage}[t]{0.35\textwidth}
    \centering
    \captionof{table}{Ablation study on the depth of the decoder. We report results on NTU 60 xsub under the linear evaluation protocol.}
  \setlength{\tabcolsep}{3.mm}{
  \begin{tabular}{l|c}
    \toprule
     \multirow{2}{*}{Depth} & NTU 60 \\
     & xsub \\
    \midrule
    2 & 84.8\\
    3 & \textbf{86.4}\\
    4 & 86.0 \\
    5 & 85.9 \\
    \bottomrule
  \end{tabular}
  \label{table: decoder depth}
  }
\end{minipage}

\end{table}
\vspace{-5pt}

\smallheading{Decoder Design} \cref{table: decoder depth} reports the effects of different decoder depths on representation learning. The best result is achieved with a depth of 3, but generally our method is robust to the decoder depth. Therefore, we adopt a depth of 5 by default considering the 
 generative capability.

\section{Conclusion}
We present MacDiff, a novel generative framework to enhance skeleton representation learning for human action understanding. By training a diffusion decoder guided by the representation from the encoder, the encoder is enforced to contain rich semantics in the representation. We formulate the objective of MacDiff as an improvement of the contrastive learning objective, theoretically demonstrating the effectiveness of the proposed framework.

\section*{Acknowledgements}
This work was supported in part by the National Natural Science Foundation of China under Grant No.62172020, and in part by the Key Laboratory of Science, Technology and Standard in Press Industry (Key Laboratory of Intelligent Press Media Technology).



%
%
\bibliographystyle{splncs04}
\bibliography{main}
\end{document}